\documentclass[runningheads]{llncs}
\usepackage{graphicx}
\usepackage{amsmath,amssymb} 
\usepackage{ruler}
\usepackage{color}
\usepackage{multirow}
\usepackage{hhline}
\usepackage{subcaption}
\usepackage{caption}
\usepackage{lipsum,widetext}

\begin{document}
\pagestyle{headings}
\mainmatter

\title{Learning Color Compatibility in Fashion Outfits} 

\author{
Heming Zhang\textsuperscript{1}, 
Xuewen Yang\textsuperscript{2}, 
Jianchao Tan\textsuperscript{3}, \\
Chi-Hao Wu\textsuperscript{1}, 
Jue Wang\textsuperscript{4}, 
C.-C. Jay Kuo\textsuperscript{1}}

\institute{
\textsuperscript{1}University of Southern California\\
\textsuperscript{2}Stony Brook University\\
\textsuperscript{3}Kwai Inc.
\textsuperscript{4}Megvii}

\maketitle

\begin{abstract}
Color compatibility is important for evaluating the compatibility of a fashion outfit, yet it was neglected in previous studies. We bring this important problem to researchers' attention and present a compatibility learning framework as solution to various fashion tasks. The framework consists of a novel way to model outfit compatibility and an innovative learning scheme. 
Specifically, we model the outfits as graphs and propose a novel graph construction to better utilize the power of graph neural networks. Then we utilize both ground-truth labels and pseudo labels to train the compatibility model in a weakly-supervised manner.
Extensive experimental results verify the importance of color compatibility alone with the effectiveness of our framework. With color information alone, our model's performance is already comparable to previous methods that use deep image features. 
Our full model combining the aforementioned contributions set the new state-of-the-art in fashion compatibility prediction.
\end{abstract}

\section{Introduction}
While many factors contribute to the compatibility of a set of fashion items, the color compatibility plays a pivotal role. When verifying whether a set of fashion items form a compatible outfit, the first check that comes to one's mind is whether their colors are compatible or not. While each human body may fit to a different silhouette, color compatibility is more universal and applicable to a wider audience. 
Learning color compatibility for fashion will help us better identify compatible outfits and provide recommendation with maximum color synergy.
Previous fashion studies largely neglected the importance of color compatibility. They either focus on the impact of each individual color \cite{[2019arxiv]Toward} or mix colors together with other information \cite{[2019SIGIR]Interpretable}. Instead, we would like to bring the importance of color compatibility to audience's attention and propose a novel solution to it.

\begin{figure}
\begin{center}
\begin{subfigure}[b]{0.3\linewidth}
    \centering
    \includegraphics[trim=0 250 500 0, clip,width=\linewidth]{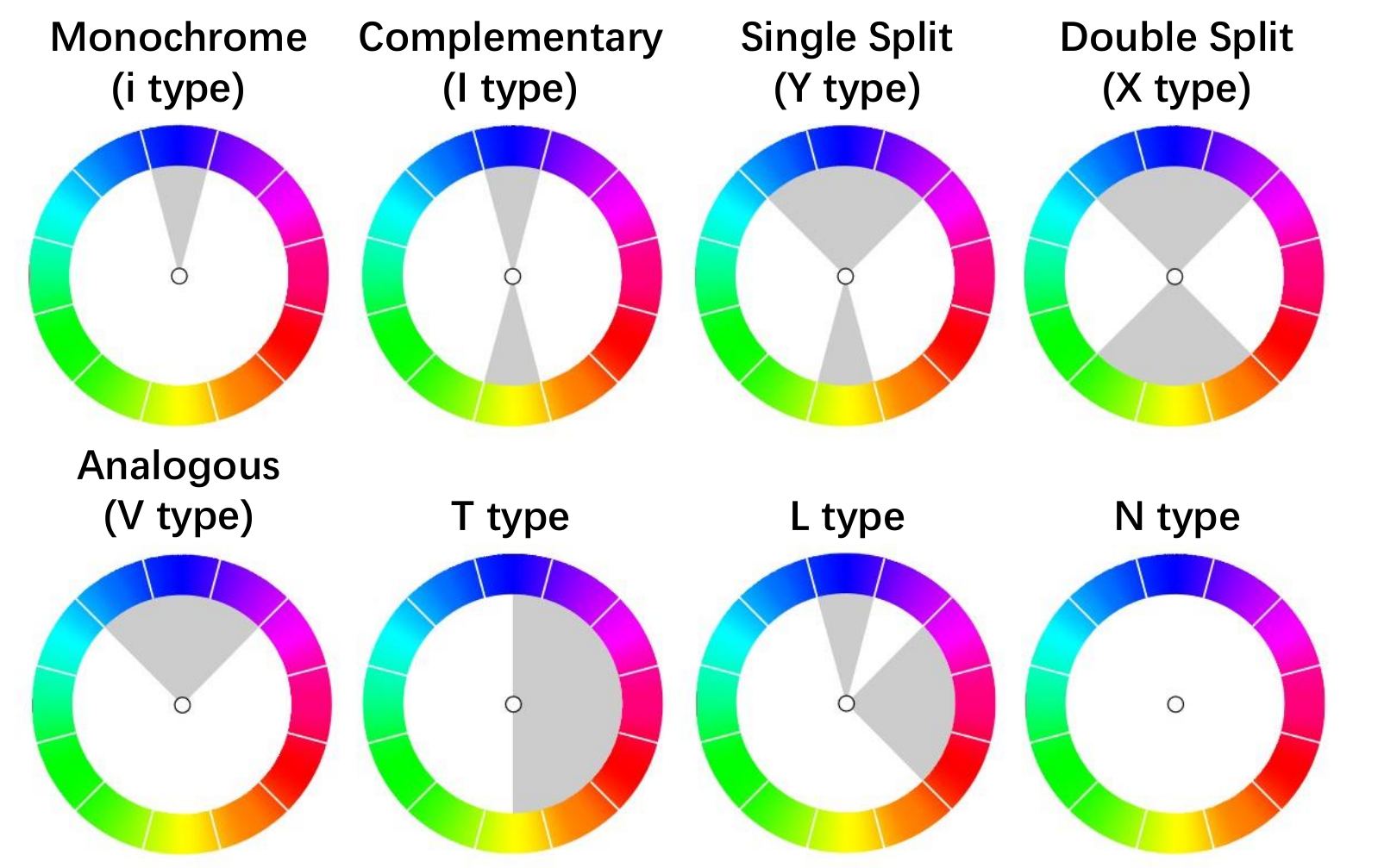}
    \caption{Previous approaches}
	\label{fig:matsuda}
\end{subfigure}
\begin{subfigure}[b]{0.65\linewidth}	
    \centering
	\includegraphics[trim=0 300 100 0, clip,width=\linewidth]{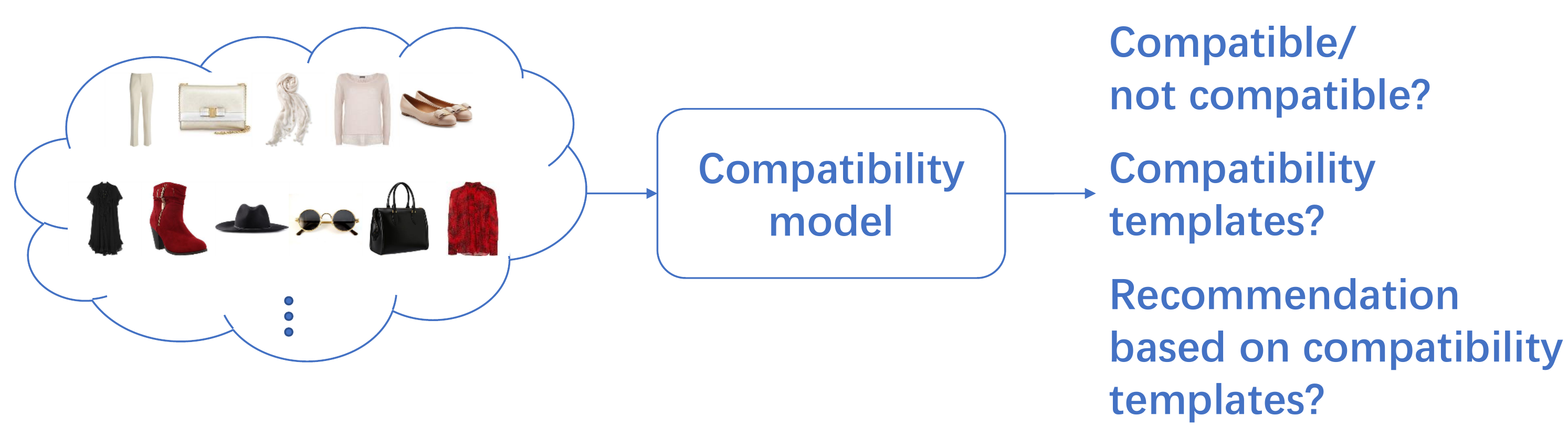}
	\caption{Our proposed approach}
	\label{fig:our_approach}
\end{subfigure}
\end{center}
   \caption{(a) Hue templates proposed by Matsuda \cite{matsuda1995color} are not designed specifically for fashion problems and deviated from real-world data. Figure modified from \cite{paletteHarmonization}.
   (b) We propose to learn color compatibility in fashion outfits from the data, which can be utilized to answer the compatibility related questions.}
\label{fig:color_compatibility}
\end{figure}

The colors of an outfit can be compatible for different reasons, e.g., similar colors versus contrasting colors. Identifying what colors are compatible in the fashion industry will help us choose compatible colors for outfits.
Therefore, instead of making a simple yes or no prediction on the compatibility of a given set of items, we would like to go one step further to learn the different templates of color compatibility in fashion.
Existing methods are not necessarily applicable to real world problems, especially to the fashion problems.
The color compatibility templates developed by Matsuda \textit{et al.} \cite{matsuda1995color,tokumaru2002color} as shown in Figure \ref{fig:matsuda} were later adopted for various purposes, e.g., natural image color harmonization \cite{cohen2006color,paletteHarmonization}. Nevertheless, O'Donovan \textit{et al.} \cite{o2011color} noticed that this theory can deviate significantly from the color data collected from human users. Moreover, those methods for natural images do not specifically apply to fashion domain. Besides, the fashion trend is notoriously volatile and changes quickly with time. Instead of using hand-weaved rules on color, it would be more preferable to learn color compatibility templates directly from fashion data.

For the reasons above, we aim to learn compatibility in fashion outfits and apply it to fashion colors for the purpose of analyzing color compatibility templates in fashion problems.
Our proposed data-driven approach for learning compatibility in fashion is illustrated in Figure \ref{fig:our_approach}. Given fashion outfit data, three consecutive problems are studied using a compatibility model: 1) compatibility prediction 2) compatibility templates extraction 3) compatibility template-based recommendation.
The compatibility model is the key to learn the compatibility from fashion outfits data. In our work, we propose a novel compatibility model, and an innovative learning scheme to train this model.

To better learn the compatibility of fashion outfits, we first model the outfits with graphs. Most previous works did not utilize graph models as they focus on learning the pairwise compatibility between two items rather than the compatibility of entire outfits \cite{[2018ECCV]Learning,[2019AAAI]TransNFCM,[2019CVPR]Context,[2019SIGIR]Interpretable,[2019TKDE]Explainable}. They obtained the outfit compatibility by simple aggregation (e.g., averaging) over all the pairwise compatibility scores. However, the pairwise compatibility does not necessarily guaranty compatibility across a entire set of items.
A few attempts have been made to model the entire outfits. A Recurrent Neural Network (RNN) was used to model the entire outfits in \cite{[2017MM]Learning}, which requires a predefined order to convert each set of items into a sequence. 

A graph representation is more suitable in modelling of fashion outfits. Each graph is orderless ensemble of nodes and their interaction (i.e. edges), thus offering a natural abstraction of the compatibility among items of a fashion outfit.
Although a graph model was used to model the entire dataset in \cite{[2019CVPR]Context}, the outfit compatibility was still obtained from pairwise item compatibility. 
Graphs are used to model outfits in the work of Cui \cite{[2019WWW]Dressing}. 
Both of them \cite{[2019CVPR]Context,[2019WWW]Dressing} took a straightforward approach of modelling each item as a node in the graphs and fed such graphs into Graph Neural Networks (GNN) or Graph Convolutional Networks (GCN). However, most GNNs/GCNs are designed for tasks such as node classification or recommendation \cite{[2017NIPS]Inductive,[2018ICLR]Few,[2018KDD]Graph}, thus focus on node information aggregation. On the contrary, in the compatibility problem one should focus on the relation among items, which is represented by the edge information in the graphs in previous studies. Such mismatch leads to the poor results in \cite{[2019CVPR]Context,[2019WWW]Dressing}, comparing to the works without graphs.
Instead of modelling fashion items as nodes, we model each pairwise relation between items as a node to utilize the full power of GNN methods to aggregate the outfit compatibility from pairwise compatibilities. This novel graph construction is more suitable for compatibility learning using GNNs/GCNs.

Popular fashion datasets only provide binary compatibility infomation, i.e. compatible or not compatible. It is not a trivial task to learn distinct compatibility templates from such binary data. To learn compatibility templates from such data, we apply a clustering algorithm to the compatible outfits to segregate outfits with distinct compatibility templates. However, a simple application of some clustering operation to the compatibility templates may not yield the segregation meaningful for human interpretation we expected. Therefore, we propose to integrate the clustering process into the compatibility learning through an iterative process, which ensures that the obtained outfit clusters are meaningful for compatibility interpretation. The obtained compatibility model is also improved, as demonstrated in the experimental results.

To summarize, our contributions are three-fold:
\begin{enumerate}
\item 
We raise the important problem of color compatibility in fashion outfits, which was largely neglected in previous work in fashion. We take a data-driven approach, which overcomes the drawbacks of previous color compatibility studies in natural images and is more applicable to the fashion domain.
\item 
We propose a new way to model the compatibility of outfits. We utilize graphs to model the entire fashion outfits with a novel graph construction method that is more suitable for compatibility learning. Our compatibility model is not limited to color compatibility, but applies to various compatibility problems.
\item 
We propose an innovative learning scheme to integrate  compatibility prediction and outfit clustering, which helps the interpretation of the fashion compatibility as well as improves the prediction performance.
\end{enumerate}

\section{Related Work}
\subsection{Color Compatibility}
Most previous studies in color compatibility are not suitable for our problem.
Color compatibility templates such as the hue templates proposed by Matsuda \textit{et al.} in \cite{matsuda1995color} and other variants were frequently used by later work \cite{tokumaru2002color,cohen2006color,paletteHarmonization}. However, experiments on large-scale color theme datasets \cite{o2011color} have demonstrated an obvious mismatch between the templates and real-world data. Instead, O’Donovan \textit{et al.} \cite{o2011color} proposed to learn a regression model for color compatibility using color theme data collected from human users. 
The limitation of this model is that it only evaluates fixed number of colors. Besides, those color themes are for general purposes and do not necessarily apply to fashion problems. 

\subsection{Fashion Outfit Compatibility}
\label{subsec:fashion_compatibility}
Most approaches only modelled pairwise compatibility between fashion items, whereas the compatibility of an entire outfit was either not considered or predicted as simple average of the pairwise compatibilities.
Some previous work used the distances of item embeddings as the measure of pairwise compatibility \cite{[2015ICCV]Learning,[2018AAAI]Dress,[2018ECCV]Learning,[2019CVPR]Complete,[2019KDD]POG,[2019WWW]Enhancing} and \cite{[2019AAAI]TransNFCM} is a generalization of this approach.
Furthermore, \cite{[2011IJCAI]Fashion,[2017MM]NeuroStylist,[2019arxiv]Toward,[2019SIGIR]Interpretable,[2019TKDE]Explainable,[2019ICCV]Learning,[2019ICIP]Learning,[2018ICMR]Interpretable,[2019arxiv]Coherent,[2019CVPR]Context,[2020CVPR]Fashion} directly modelled the pairwise compatibility. Jagadeesh \textit{et al.} \cite{[2014KDD]Large} used color representations as clues to retrieve fashion items that are matched to the query.

It is challenging to model the compatibility of entire outfits with various number of items. In \cite{[2017TMM]Mining}, item features were fused by global pooling. 
Similarly, in \cite{[2017ICCVworkshop2018WACV]}, items are divided into six categories and the six features are concatenated.
RNNs were utilized in \cite{[2017MM]Learning,[2018KDDworkshop]Outfit}, but a pre-defined order is required to model a set of items as a sequence.
Although \cite{[2019CVPR]Context} directly modelled the pairwise compatibility, context information from other items were used. The entire dataset is modelled as a graph, where each node represents an item and nodes that appear in the same outfit are connected. During inference, the neighbors of each node contains items that appear in other outfits and provide rich context information. Such setting is different from previous work where each outfit is evaluated separately and no information from other outfits is provided. Experiments indicate that the method in \cite{[2019CVPR]Context} performs poorly in the common setting.

Closely related to our work, Cui\cite{[2019WWW]Dressing} proposed to model each outfit as a graph where each node represents an item with a specific category. However, the proposed GNN model \cite{[2019WWW]Dressing} learned on such graphs focuses more on item category information and neglected the relation among items. Instead, we propose a novel graph construction that is more suitable for the compatibility problem. 

\subsection{Deep Learning on Graphs}
Graph neural networks (GNN) and graph convolutional networks (GCN) have been evolved over the years \cite{bruna2013spectral,henaff2015deep,defferrard2016convolutional,kipf2016semi,[2017NIPS]Inductive,[2018ICLR]Graph}. A common drawback of the existing models is that they only utilized the node features. In their work, edge features were either ignored or only the 1-D edge weights were used. 

A few attempts were made to overcome this drawback. In \cite{[2017CVPR]Dynamic}, an edge-conditioned convolution was proposed where edge information was utilized to filter the nodes. Moreover, edge features were directly utilized for node feature aggregation functions in \cite{[2018arxiv]Exploiting}. In these works, edge features are only considered as an auxiliary for node feature aggregations and do not get updated throughout the network inference.
Monti et al. \cite{[2018arxiv]Dual} proposed a Dual-Primal GCN framework, which alternated between the original graph (primal) and its line graph (dual). The dual convolutional layer produces features on the edges of the primal graph, and the primal edge features were used in the primal convolutional layer to produce primal node features. In \cite{[2019CVPR]Edge}, 2-D edge features defined as (dis)similarities between pairs of nodes were updated following each update of nodes. Five options for updating edge features were discussed in \cite{[2019TOG]Dynamic}. The edge features in these works are only used for node feature updates. For fashion compatibility, however, direct aggregation and update of edge features are desired.

\section{Learning Color Compatibility in Fashion Outfits}
\label{sec:method}

The fashion compatibility prediction problem takes a set of fashion items $\mathbf{x}$ as inputs and a binary compatibility label $y$ as the prediction target. For each fashion item, an image is provided, and additional information such as category annotations or text descriptions may also be available. 

We formulate the compatibility template learning problem as a clustering problem. 
The compatible outfits are grouped into clusters, where each cluster obtained should identify an outfit compatibility template.

\begin{figure*}
\begin{center}
	\includegraphics[trim=0 380 370 0, clip,width=1\linewidth]{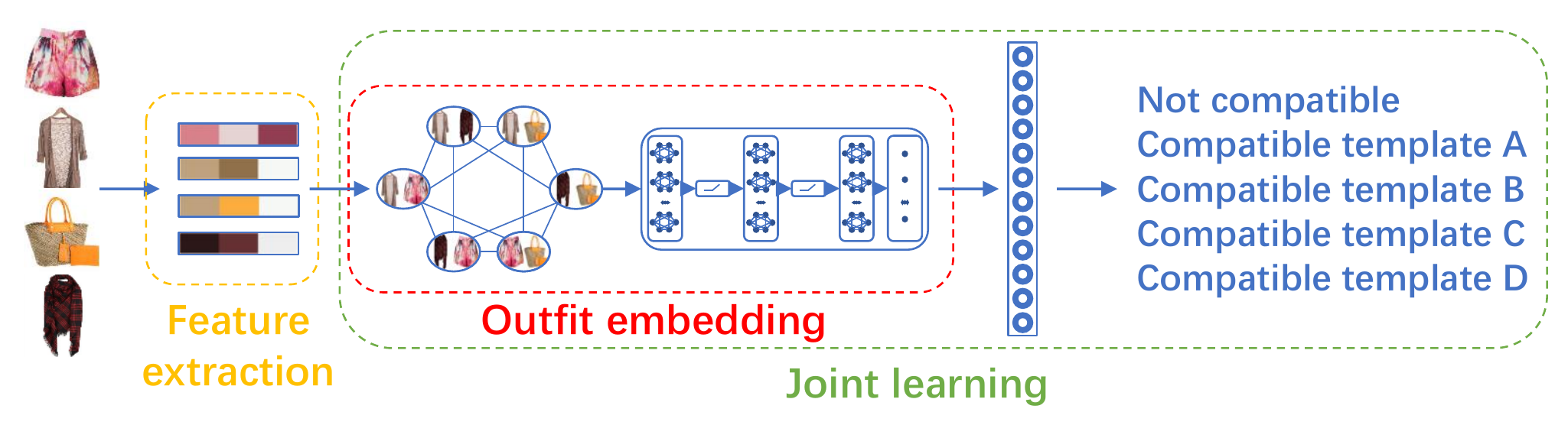}
\end{center}
   \caption{An overview of our proposed method for learning color compatibility in fashion outfits.}
\label{fig:overview}
\end{figure*}

An overview of our color compatibility learning framework is shown in Figure \ref{fig:overview}. We first extract color features from the items, which will be described in Section \ref{subsec:palette}. As introduced in Section \ref{subsec:graph}, we then embed the outfit into an embedding feature, which will be used for compatibility prediction. We will explain our proposed learning scheme for the compatibility model in Section \ref{subsec:joint}.

\subsection{Feature Extraction}
\label{subsec:palette}
For the problem of color compatibility learning, we propose to use a color palette, collection of dominant colors, as the color feature for an item. Previous work \cite{[2013ICCV]Paper,[2014MM]What,[2015MM]Collaborative} on fashion study usually use color histogram as features. Here we use the color palette instead for several reasons: First, our goal is to study the compatibility among colors, rather than the statistics of color distribution; Second, noises such as illumination variation, are encoded in the color histogram but color palette is more invariant to such variations; Third, color palettes are more abstractive, from which we can better learn the compatibility templates.

To extract a color palette from a fashion item image, we convert the image color space from $\mathit{RGB}$ to $\mathit{Lab}$.
In the $\mathit{Lab}$ color space, the distance between two colors indicates their visual difference in humans perception. For each image, we apply the K-means algorithm to cluster the pixels with their Lab colors as features. Each cluster centroid represent a dominant color in the image and the cluster centroids together form the palette of the corresponding fashion item. The color feature of the image is obtained by concatenating the cluster centroids. In our work, we cluster the colors into 3 clusters. The resulting color feature has 9 elements (the details are in the supplemental material).

Although we use color features for the purpose of color compatibility learning, our proposed compatibility learning framework can be applied to other representations as well. In section \ref{sec:experiment}, we demonstrate the effectiveness of our framework with other fashion item representations such as image features extracted from ResNet \cite{resnet}.

\subsection{Outfit Embedding}
\label{subsec:graph}
Our outfit embedding consists of two steps. We first construct an outfit graph to model the given set of items, then embed the outfit graph into an outfit embedding.

As explained in previous sections, we would like to overcome the mismatch between outfit graphs and existing GNN models. Rather than designing a network structure specific to our problem, we choose to design a new graph construction method that fits to most existing models that aggregate the node information. Since the outfit compatibility is the global relation among outfit items, we propose to use nodes to represent pairwise item relations instead of individual items. Thus the global outfit relation can be aggregated from local pairwise relations. 

\begin{figure}[t]
\begin{center}
   \includegraphics[trim=0 190 310 0, clip,width=0.8\linewidth]{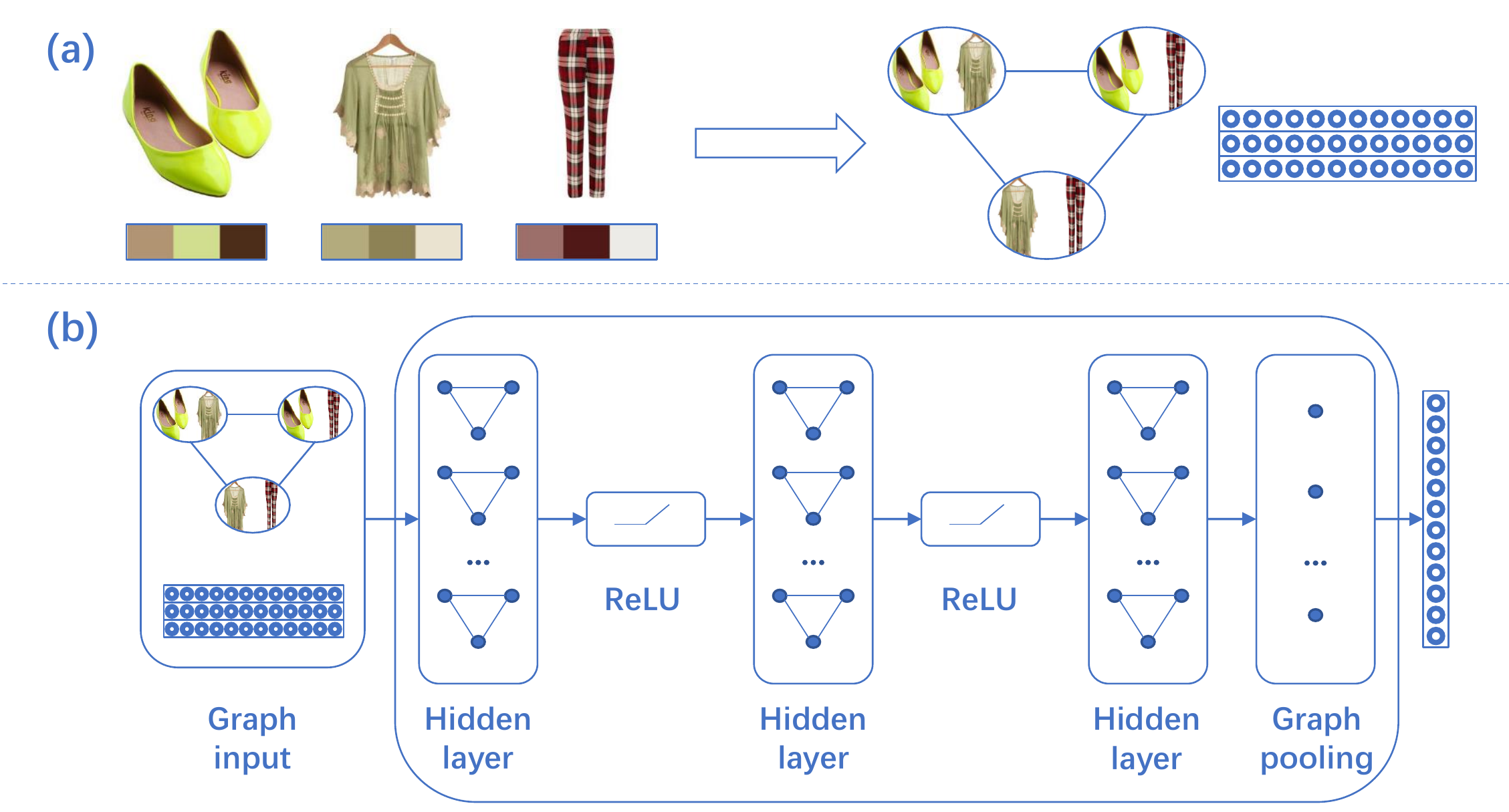}
\end{center}
   \caption{(a) The graph construction step by modelling each pairwise item relation as a node. The node features are obtained by embedding two item features. (b) The graph embedding step with GNN.}
\label{fig:graph}
\end{figure}

For a given outfit of $N$ items, we construct a graph $G=(V,E)$ with $\binom{N}{2}$ nodes where each node represents the pairwise relation between two items in the outfit. For each pair of nodes with shared item, we connect them with an edge. An illustration is shown in Figure \ref{fig:graph}. Comparing to the graph construction method in \cite{[2019WWW]Dressing}, our proposed graph can be viewed as its line graph (also known as the edge-to-node dual graph). 

For the node $\mathcal{V}_{(i,j)} \in V$ that represents the relation between the $i$th and the $j$th items, the node feature is obtained from features of these two items as
\begin{equation}
f_{\mathcal{V}_{(i,j)}} = Z(f_i) \odot Z(f_j),
\end{equation}
where $Z(\cdot)$ is an embedding module, $\odot$ denotes the Hadamard product and $f$ denotes the feature.

For an outfit graph constructed from an outfit, we then embed the outfit graph into an embedding of length $d$, where $d$ is a hyper-parameter. 
We can apply various models to aggregate the node information and embed the graph. In this work, we use GraphSage \cite{[2017NIPS]Inductive} for demonstration. At $l$-th layer, the hidden representation of node $\mathcal{V}_{(i,j)}$ in $G$ is updated as
\begin{equation}
\begin{split}
m_{\mathcal{V}_{(i,j)}}^{l+1} &= \text{MEAN} \left(\{h_{\mathcal{V}'}^l, \forall  \mathcal{V}' \in \mathcal{N}(\mathcal{V}_{(i,j)}) \} \right), \\
h_{\mathcal{V}_{(i,j)}}^{l+1} &= \sigma \left(W^l [h_{\mathcal{V}_{(i,j)}}^l, m_{\mathcal{V}_{(i,j)}}^{l+1}]+b\right), 
\end{split}
\end{equation}
where $\mathcal{N}_{\mathcal{V}_{(i,j)}}$ is the neighborhood of $\mathcal{V}_{(i,j)}$, $[\cdot,\cdot]$ denotes concatenation, $W$ is the weight matrix, $b$ is the bias term and $\sigma$ is a nonlinear activation function. We first obtain the new message $m_{\mathcal{V}_{(i,j)}}^{l+1}$ by aggregating information from the neighbors of node $\mathcal{V}_{(i,j)}$, then the updated hidden representation $h_{\mathcal{V}_{(i,j)}}^{l+1}$ is obtained from the new message and the previous hidden representation of node $\mathcal{V}_{(i,j)}$. The node features at the last layer are then globally pooled into a graph embedding. 

Given the graph embedding, we can apply a fully connected layer and take a softmax to predict the compatibility from the outfit graph embedding. To train this layer together with the GNN model, we propose a novel learning scheme.

\subsection{Joint Learning Scheme}
\label{subsec:joint}
There are different reasons for a set of colors to be considered compatible, e.g. similar colors or different colors with a synergy. To further understand such differences and explore compatibility templates, we cluster the compatible outfits in the embedding space such that the outfits in the same cluster should be close in the embedding space. Note that the incompatible samples are all simply assigned to a single ``incompatible" cluster. We do not care why a set is not compatible so there is no need of finer clustering for them. (Detailed discussion of a single incompatible cluster is in the supplemental material).

We expect each outfit cluster to represent a distinct and tangible reason for the containing outfits to be considered compatible. Therefore, each cluster can be interpreted as a compatibility template. However, it is not trivial to cluster the outfits such that compatibility templates could be interpreted.
If we first train the embedding with outfit compatibility labels and then cluster the outfits using their trained embeddings, the clusters obtained may not be meaningful for compatibility prediction and thus may not help us to interpret compatibility templates.

Therefore, we propose a joint learning scheme that brings the clustering process into the training loop of the compatibility model. It is an iterative learning scheme that alternatively generates pseudo labels from the outfit clusters and updates the network with pseudo labels. 

\subsubsection{Pseudo label generation.} 
Originally a sample $i$ contains the outfit input $\mathbf{x}_i$ and a binary outfit compatibility label $y_i \in \{0,1\}$. For each sample $i$, we generate a multi-class pseudo label $\tilde{y}_i=c$, where $c \in \{0,1,\cdots,C\}$ and $C$ is the number of compatible clusters.
The pseudo label generation procedure consists of four steps and is illustrated in Figure \ref{fig:joint}:

\begin{figure}[t]
\begin{center}
   \includegraphics[trim=0 250 350 0, clip,width=0.8\linewidth]{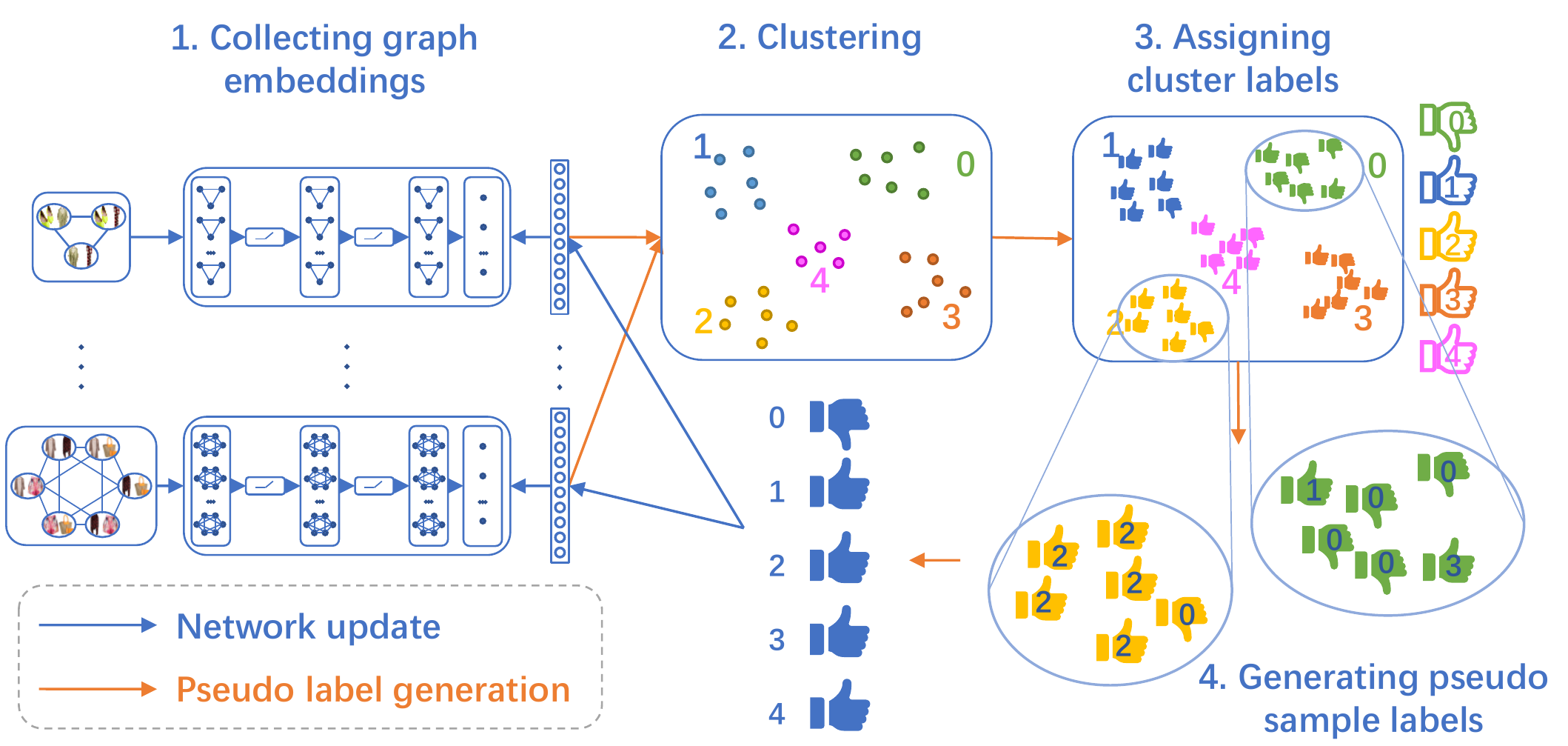}
\end{center}
   \caption{Illustration of the proposed joint compatibility prediction and outfit clustering method. We alternatively generate pseudo labels for every sample and update the network parameters using the pseudo labels. We show five clusters for illustration. The cluster 0 is assigned with negative label, whereas clusters 1-4 are assigned with positive labels basing on their compatible sample ratios.}
\label{fig:joint}
\end{figure}

\textbf{1. Collecting graph embeddings}. We collect the outfit embeddings of all outfit samples in the training set using the current model.

\textbf{2. Clustering}. The current embeddings of outfits are clustered to $C+1$ clusters and the clusters are indexed from 0 to C. For cluster $c \in \{0,1,\cdots, C\}$, we count the number of compatible outfits $m_{c,pos}$ and the number of incompatible outfits $m_{c,neg}$ it contains. Then we calculate the compatible ratio as $r_c = m_{c,pos} / m_{c,neg}$ for each cluster $c$.

\textbf{3. Assigning cluster labels}. In this step, we assign a cluster label $z_c \in \{0,1\}$ to cluster $c \in \{0,1,\cdots,C\}$.  These labels are only assigned to the clusters and do not represent for the outfit sample they contain.
The obtained clusters are ranked by the ratios $r_c$ obtained in the previous step. 
Then the top $C$ clusters are assigned with cluster labels of 1, whereas the rest cluster is assigned with a cluster label of 0. 
 
\textbf{4. Generating pseudo sample labels}. For each data sample $(\mathbf{x}_i,y_i)$, we assign a pseudo multi-class label $\tilde{y}_i \in {0,1,\cdots,C}$. The Euclidean distance between the $i$th sample and the centroid of cluster $c$ in the embedding is denoted as $d(i,c)$. The pseudo label is assigned as
\begin{equation}
\begin{split}
    \tilde{y}_i = \underset{c}{\arg\min} \; d(i,c)  \quad
    \text{s.t.} \; y_i = z_c,
\end{split}
\end{equation}
which is the index of the closest cluster that has the same label.

The unsupervised learning method DeepCluster \cite{[2018ECCV]Deep} has a similar concept which
iteratively clusters the features obtained from the network, and update the network using the clusters. In our method, we make use of the available compatibility labels and obtain the pseudo labels by fusing the ground-truth compatibility labels with the cluster assignments.

\subsubsection{Network update with pseudo labels.}
We combine two loss functions to update the network with both the pseudo and ground-truth labels. 
For each compatible outfit, we randomly sample an incompatible outfit to pair with it.


The first loss is a global loss that is applied to entire outfits. We use the categorical cross-entropy loss to update the network parameters with pseudo labels: 
\begin{equation}
\begin{split}
l_1 =\sum_i  [l_{CCE}(\left[g(\mathbf{x}_{i,pos})\right]_{\tilde{y}_{i,pos}}, \, \tilde{y}_{i,pos}) 
+ l_{CCE}(\left[g(\mathbf{x}_{i,neg})\right]_{\tilde{y}_{i,neg}}, \, \tilde{y}_{i,neg})],
\end{split}
\label{eq:l1}
\end{equation}
where $l_{CCE}(\cdot,\cdot)$ is the categorical cross-entropy loss, $g[\cdot]_c$ is the predicted softmax score for the class $c \in \{0,1,\cdots, C\}$, and the subscriptions $_{i,pos}$ and $_{i,neg}$ denote the positive and the negative outfits of the $i$th pair of outfits, respectively. 

Additionally, if a positive/negative pair of outfits share common items, we apply a local loss to subsets of the outfits. It helps the network learn more subtle differences between item relations. For each positive/negative pair of outfits that share items, we remove their common items from both outfits and obtain a pair of outfit subsets.  Then for this pair of outfit subsets, we apply a binary cross-entropy loss with the compatibility labels:
\begin{equation}
\begin{split}
l_2 =  \sum_i  [l_{BCE}(g(\hat{\mathbf{x}}_{i,pos}),\, y_{i,pos}) 
+ l_{BCE}(g(\hat{\mathbf{x}}_{i,neg}),\, y_{i,neg})], 
\end{split}
\label{eq:l2}
\end{equation}
where $l_{BCE}(\cdot,\cdot)$ denotes the binary cross-entropy loss and $\hat{\mathbf{x}}$ denotes the subset of $\mathbf{x}$ obtained by removing the common items. The intuition is that a subset of a compatible outfit should also be compatible, and a subset of an incompatible outfit is most likely to be incompatible.

The final loss function takes the weighted sum of $l_1$ and $l_2$ with a hyper-parameter $\lambda$
\begin{equation}
l = l_1 + \lambda l_2.
\label{eq:loss}
\end{equation}

More details of the loss functions are in the supplemental material.
\section{Experimental Analysis}
\label{sec:experiment}

\subsection{Datasets}
\textbf{Polyvore Outfits} \cite{[2018ECCV]Learning}. It contains 53,306, 5000, and 10,000 outfits for training, validation, and testing, respectively. It provides one image per fashion item and some items also come with text descriptions. 
Two tasks are provided for benchmarking. One task is the compatibility prediction (CP), in which the compatibility predicted for given outfits are evaluated using area-under-curve (AUC). The other task is the fill-in-the-blank (FITB), where one fashion item out of four candidates that fits a given outfit subset best is chosen and the evalutation metric is the accuracy.

\textbf{Maryland Polyvore} \cite{[2017MM]Learning}. This dataset is similar to the Polyvore Outfits dataset, but much smaller in size. It has 17,316, 1,407, and 3076 outfits for training, validation and testing, respectively. It also has the two tasks (CP and FITB) for benchmarking. The test set provided by \cite{[2017MM]Learning} is not challenging enough as it contains negative samples that can be easily identified using category information. We follow the previous studies \cite{[2018ECCV]Learning,[2019ICCV]Learning} and evaluate our model on a more challenging test set proposed in \cite{[2018ECCV]Learning}.

\subsection{Implementation Details}

At the preprocessing step, we first separate fashion items from image backgrounds using \cite{opencv}. Then we extract a 3-color palette from each fashion item.

We use one fully connected layer for node feature embedding and a 3-layer GraphSage \cite{[2017NIPS]Inductive} with ReLU activation and global max-pooling for graph embedding. The hidden representations are 60-D and the final embedding is 20-D. We utilize K-means with cosine similarity for clustering and re-assign the pseudo labels to outfits in every other 25 epochs of training. We tuned the hyper-parameters and chose $C=4$ and $\lambda=0.5$ for the best performance. Detailed analysis can be found in the supplemental material.

Since our compatibility model also applies to other representations, we follow \cite{[2018ECCV]Learning,[2019ICCV]Learning} and extract image features from ResNet18 \cite{resnet}, as well as text features from HGLMM Fisher vectors \cite{hglmm} of word2vec \cite{word2vec} in some experiments. 

\subsection{Comparison with Previous Work}
\begin{table*}[!t]
\begin{center}
\caption{Comparison with previous methods using color palettes or deep image features.}
\label{tb:img_results}
\begin{tabular}{|p{2.6cm}|p{2cm}|c|p{0.8cm}|p{0.8cm}|p{0.8cm}|p{0.8cm}|}
\hline
\multirow{2}{2.6cm}{Methods}  & \multirow{2}{2cm}{Image representation} & \multirow{2}{1.2cm}{Needs category}
& \multicolumn{2}{|c|}{P. Outfits} & \multicolumn{2}{|c|}{Maryland P.} \\
\cline{4-7}
& & & CP & FITB & CP & FITB \\
\hhline{|=|=|=|=|=|=|=|}
Siamese Net\cite{[2015ICCV]Learning} & ResNet18 & - &0.81 & 52.9 & 0.85 & 54.4\\
\hline
Type-aware\cite{[2018ECCV]Learning} & ResNet18 & Yes & 0.83 & 54.0 & 0.87 & 57.9\\
\hline
Outfit gen.\cite{[2019arxiv]Coherent} & ResNet18 & - &
0.90 & 59.1 & 0.89 & 59.8 \\
\hline
Context-aware\cite{[2019CVPR]Context} & ResNet50 & - &
--- & --- & 0.76 & 47.0 \\
\hline
NGNN\cite{[2019WWW]Dressing} & ResNet18 & Yes & 
0.78 & 53.0 & 0.66 & 37.2\\
\hline
CSA-Net\cite{[2020CVPR]Fashion} & ResNet18 & Yes &
\textbf{0.91} & \textbf{63.7} & --- & ---\\
\hline
\multirow{2}{2.6cm}{Ours} & Color palette & - & 
0.84 & 58.0 & 0.83 & 49.1\\
& ResNet18 & - &
\textbf{0.91} & 59.2 & \textbf{0.91} & \textbf{59.9}\\
\hline
\end{tabular}
\end{center}

\end{table*}

We compare with the results of the following methods:

\textbf{Bi-LSTM} \cite{[2017MM]Learning}. It models the outfits as sequences and the orders are pre-defined according to item categories. The image features are obtained from Inception V3 \cite{inceptionv3}. It utilizes the text descriptions as regularization during training.

\textbf{Siamese Network} \cite{[2015ICCV]Learning}. It models the pairwise compatibility. We compare with its results reported in \cite{[2018ECCV]Learning}.

\textbf{Type-aware} \cite{[2018ECCV]Learning}. 
It considers the category differences when learning the pairwise compatibility, which additionally requires the item category annotation. Same as Bi-LSTM, the text features are used as regularization during training.

\textbf{Outfit generation} \cite{[2019arxiv]Coherent}. It learns the pairwise compatibility. BERT features \cite{bert} are extracted from text descriptions and used for both training and testing.

\textbf{Context-aware} \cite{[2019CVPR]Context}. It models the entire dataset as a graph. Its best results are obtained by utilizing context information from other outfits, which is a different setting from most previous works. We only compare to its results obtained without extra context information from other outfits. Its image features are extracted from ResNet50 \cite{resnet}.

\textbf{NGNN} \cite{[2019WWW]Dressing}. It models each outfit as an graph where each item is modelled as a node.
Item category annotation is required for graph construction. The authors only reported the results on the easy test set of the small Maryland Polyvore dataset. We conduct the experiments using their original source code on the challenging test sets provided by \cite{[2018ECCV]Learning}. 

\textbf{SCE-Net} \cite{[2019ICCV]Learning}. It models pairwise compatibility and requires text features for both training and testing.

\textbf{CSA-Net} \cite{[2020CVPR]Fashion}. It models pairwise compatibility and its category-based subspace attention network requires item category annotation.

\textbf{Ours}. We extract color features as described in Section \ref{subsec:palette}. To further verify the effectiveness of our compatibility system, we also utilize ResNet18 features and HGLMM features following most previous works. Our framework does not require item category annotation and the text features are only used for training as regularization.

The comparisons are conducted in two groups. In the first group, the images are used as inputs for both training and testing. The image representations used are either color palettes extracted using our method, or deep image features extracted from convolutional neural networks. Among the first group, Type-Aware\cite{[2018ECCV]Learning}, NGNN\cite{[2019WWW]Dressing} and CSA-Net\cite{[2020CVPR]Fashion} also requires item categories as inputs.
The results of the first group are listed in Table \ref{tb:img_results}. With color palettes of 3 colors as image representation alone, our results already outperform most previous methods with deep features on the large-scale Polyvore Outfits dataset. This observation verifies our claim of the importance of color compatibility. Using the same deep features further boosts our results. Only the CSA-Net, which requires extra category annotation, outperforms ours. It demonstrates the effectiveness of the proposed compatibility learning method. 
The other graph method NGNN suffers from the bias on item categories. Its performance on the Maryland Polyvore drop 30\%-40\% on both tasks from the easy test set to the hard test set. Our graph method better utilizes individual item characteristics.

\begin{table*}[!t]
\begin{center}
\caption{Comparison with previous methods when text information is additionally utilized.}
\label{tb:img_text_results}
\begin{tabular}{|p{2.05cm}|p{3.15cm}|c|c|p{0.8cm}|p{0.8cm}|p{0.8cm}|p{0.8cm}|}
\hline
\multirow{2}{2.05cm}{Methods} & \multirow{2}{3.15cm}{Image / text representation} & \multirow{2}{1.2cm}{Needs category} 
& \multirow{2}{1.55cm}{Needs text in testing}
& \multicolumn{2}{|c|}{P. Outfits} & \multicolumn{2}{|c|}{Maryland P.} \\
\cline{5-8}
& & & & CP & FITB & CP & FITB \\
\hhline{|=|=|=|=|=|=|=|=|}
Bi-LSTM\cite{[2017MM]Learning} & InceptionV3/HGLMM & Yes & - &
0.65 & 39.7 & \textbf{0.94} & 64.9\\
\hline
Type-aware\cite{[2018ECCV]Learning} & ResNet18/HGLMM & Yes & - & 
0.86 & 55.3 & 0.90 & 61.0\\
\hline
Outfit gen.\cite{[2019arxiv]Coherent} & ResNet18/BERT & - & Yes &
\textbf{0.93} & \textbf{66.1} & 0.93 & \textbf{69.9}\\
\hline
SCE-Net\cite{[2019ICCV]Learning} & ResNet18/HGLMM & - & Yes &
0.91 & 61.6 & 0.90 & 60.8\\
\hline
Ours & ResNet18/HGLMM & - & - & 
0.90 & 65.8 & 0.91 & 67.1\\
\hline
\end{tabular}
\end{center}
\end{table*}


In the second group, the text descriptions are also used for the network training. Unlike Bi-LSTM, Type-aware and ours, Outfit gen. and SCE-Net also need text descriptions during testing. 
Similar conclusion can be drawn from the results of the second group listed in Table \ref{tb:img_text_results}.
When switching from the small-scale Maryland Polyvore dataset to the large-scale Polyvore Outfits dataset, the performances of Bi-LSTM and Type-aware drop dramatically, indicating potential overfitting. On the contrary, the proposed method gives consistent good results on both datasets. 
Although SCE-Net utilizes text description during inference,  our method still shows comparable results on the CP task, while improving the FITB task by a significant margin of 4\%-6\%. Our results are only outperformed by Outfit gen., which uses a more powerful text features both for training and inference. Although the text descriptions provide more useful information during inference, it is less applicable to real scenarios where text descriptions are unavailable.

\subsection{Ablation Studies}
\label{subsec:ablation}
We compare the following modules as ablation studies to evaluate the contribution of each proposed module:
NG denotes the node graph where each node represents an item in the outfit. LG denotes our proposed line graph where each node represents a pairwise relation between items. 
$l_{BCE}$ is the binary cross-entropy loss. $l_1$ and $l_2$ are the two losses proposed in (\ref{eq:l1}) and (\ref{eq:l2}), respectively.
The ablation results are evaluated on the validation set of the large-scale Polyvore Outfits dataset and listed in Table \ref{tb:ablation}. All methods compared use color palettes.
\begin{table}
\begin{center}
\begin{tabular}{|l|c|c|}
\hline
Methods & 
Compat AUC & FITB Acc\\
\hhline{|=|=|=|}
NG+$l_{BCE}$& 0.76 & 42.0\\ 
LG+$l_{BCE}$ & 0.84 & 52.7 \\
LG+$l_1$ &  0.84 & 54.7 \\
LG+$l_1$+$l_2$ &  0.84 & 58.8\\
\hline
\end{tabular}
\end{center}
\caption{Ablations studies on how each proposed module contributes to the final performance. The experiments are conducted on the validation set of Polyvore Outfits with color palettes as fashion representations.}
\label{tb:ablation}
\end{table}


Each proposed module contributes to the final performance. 
Replacing the node graph with our proposed line graph, the overall performance is dramatically improved, supporting our argument that our graph better represents the fashion outfits for compatibility learning.
By replacing the ground-truth binary labels to generated multi-class pseudo labels, the accuracy of the FITB task is increased by 2\%. It shows that incorporating outfit clustering into the training loop not only provides us meaningful clusters but also helps the model better understand the compatibility.
With the entire joint learning scheme, the accuracy of the FITB task is boosted by 6\%. It verifies that our joint learning scheme helps the network learn the subtle difference of item relations. 

\begin{figure*}[!th]
\begin{center}
\begin{subfigure}[b]{\linewidth}
	\centering
	\includegraphics[trim=0 400 300 0, clip,width=\linewidth]{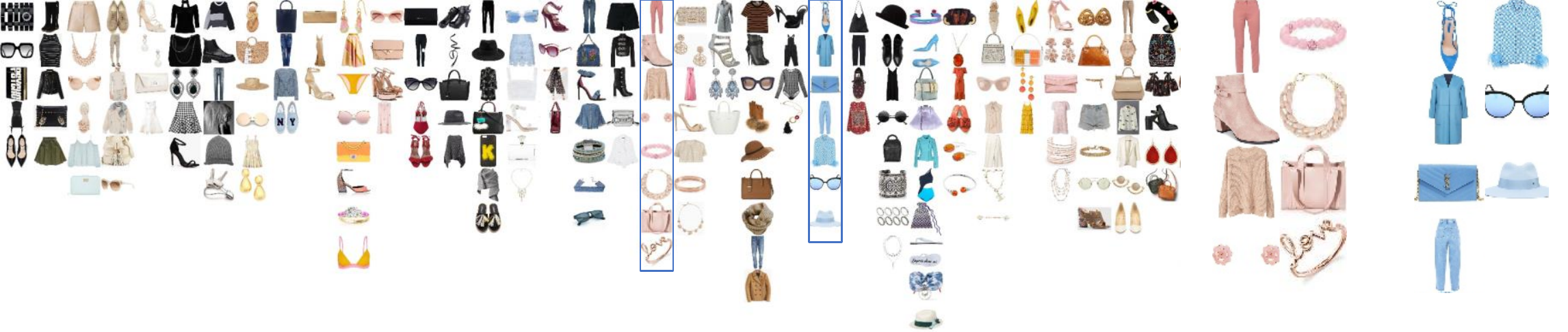}
	\caption{Cluster 1: The visualization suggests that the items of the outfits in this cluster have similar colors.}
	\label{fig:sample1}
\end{subfigure}
\begin{subfigure}[b]{\linewidth}
	\centering
	\includegraphics[trim=0 400 300 0, clip,width=\linewidth]{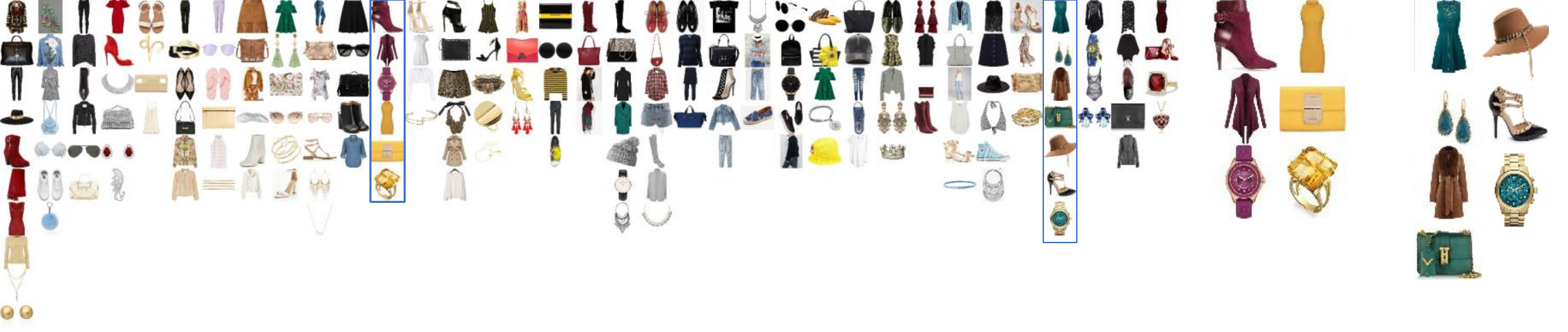}
	\caption{Cluster 2: The visualization suggests that the outfits in this cluster have the combination of two major colors that distinct besides of the neutral colors.}
	\label{fig:sample2}
\end{subfigure}
\end{center}
   \caption{Compatible outfit clusters predicted using color palettes as fashion representation. Due to limited space, we only display 35 outfits per cluster (one outfit per column) and at most 10 items per outfit. More results can be found in supplementary materials. Two outfit samples in the bounding boxes are enlarged and shown on the right.}
\label{fig:sample_color}
\end{figure*}
\subsection{Outfit Cluster Visualization}
\label{subsec:cluster_viz}
To gain more insights of our color compatibility model, we visualize the compatible clusters obtained from learning the fashion outfits data. For each outfit sample in the validation set of the Polyvore Outfits dataset, we obtain its predicted cluster using our model. Due to the limited space, only two clusters with 35 outfit samples per cluster are displayed in Figure \ref{fig:sample_color}. More visualization results are found in the supplemental material. For each cluster, two outfit samples are enlarged and displayed on the right.

\begin{figure*}[!th]
\begin{center}
\includegraphics[trim=0 195 0 0, clip,width=\linewidth]{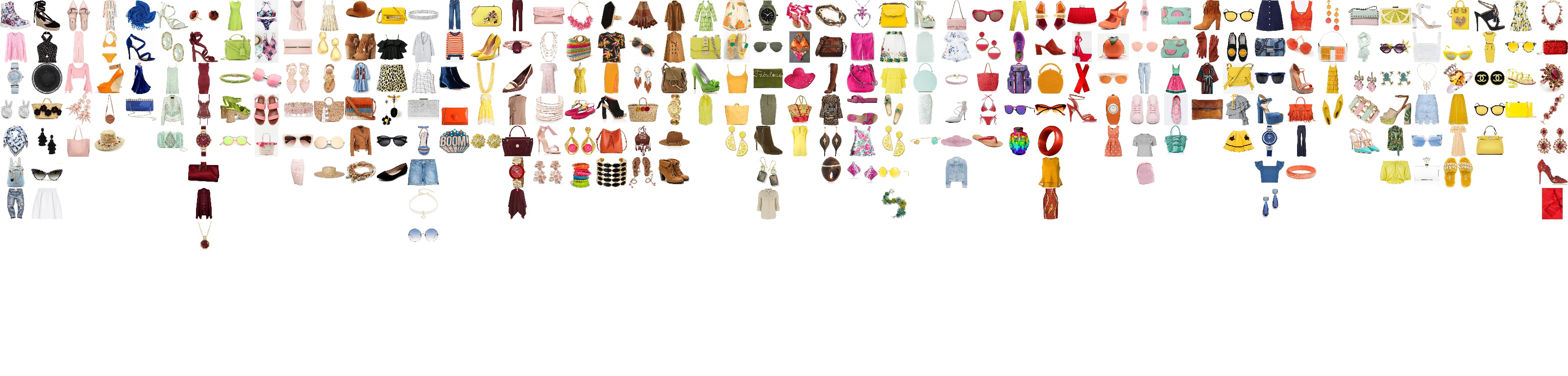}
\end{center}
\caption{A compatible outfit cluster predicted using deep image features that reveals a color template. The items in the same column belong to the same outfit. The visualization suggests that the items of the outfits in this cluster have similar color.}
\label{fig:sample_it}
\end{figure*}

Each cluster has its own color compatibility template. The visualization in Figure \ref{fig:sample1} suggests that the items in each outfit share similar colors. The cluster in Figure \ref{fig:sample2} demonstrates the combination of two major colors that are distinct besides of the neutral colors with low chroma such as black or white. 
As illustrated via the visualization, our model does not simply group outfits together that share the similar colors, such as one cluster with green colors and the other with yellow colors. The outfits sharing similar color compatibility templates are grouped together through our model, which is the purpose of integrating the clustering process into the training loop.

It is not surprising that we can also find some color compatibility templates in the visualization of the model trained on deep image features. One example of the clusters is shown in Figure \ref{fig:sample_it}, which displays the color compatibility template of similar colors. It again verifies our claim that color is an essential impact factor for outfit compatibility.

\begin{figure}[!th]
\begin{center}
\includegraphics[trim=0 280 500 0, clip,width=\linewidth]{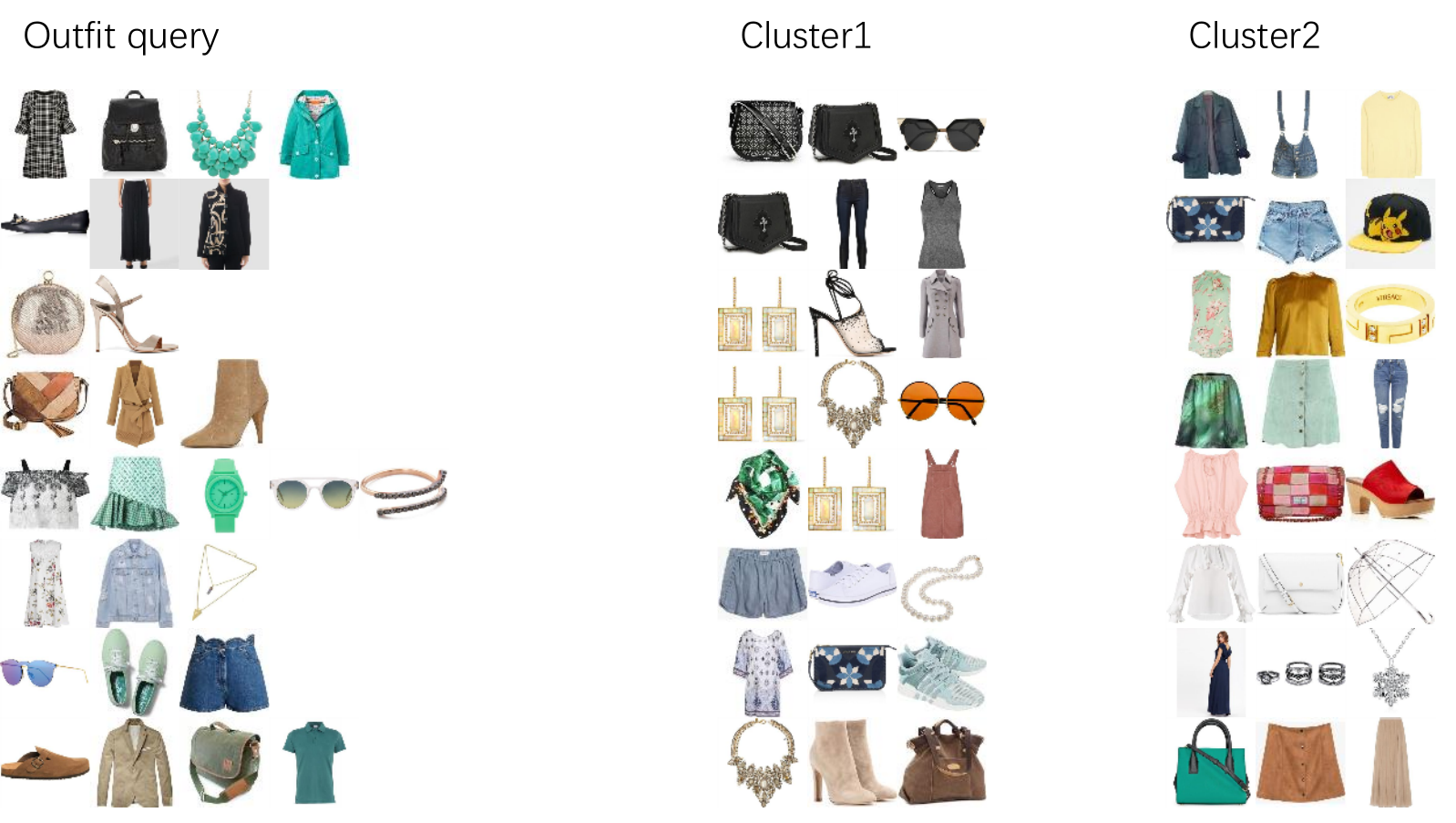}
\caption{Recommendations based on different color templates obtained from the outfit clusters.}
\label{fig:recommendation12}
\end{center}
\end{figure}

\subsection{Fashion Recommendation based on Color compatibility Templates}
While we can select one single overall best fit, as with the existing methods. Our innovative clustering method enables more versatile, customized recommendations from different compatibility templates. 
We made use of the outfit queries from the FITB task in the Polyvore Outfits dataset, and randomly sampled 2000 items as candidates. For each outfit query, we compare all the candidates and generate recommendation based on the cluster prediction. 
Figure \ref{fig:recommendation12} shows the recommendation from clusters 1 and 2. More results are in the supplemental material. For each outfit query, three recommendations per cluster are displayed.
When the query outfit has similar colors, recommendation from cluster 1 also suggests a similar color, whereas cluster 2 suggests a distinct color. When the query outfit has mixed colors, cluster 1 matches one of those colors, whereas cluster 2 either matches one of those colors or suggests a neutral color.

\section{Conclusion}
In this paper, we addressed an important problem, namely learning color compatibility in fashion outfits. It plays an essential role in outfit compatibility, yet it has been neglected in previous work. We proposed a compatibility learning method including a novel compatibility model and an innovative learning scheme. Our approach provides more insights and possibility for customized recommendation.
The proposed framework is not only suitable for color compatibility learning task, but also extendable to general outfit compatibility learning tasks. 
We conducted comprehensive experiments to demonstrate the importance of color compatibility, and the effectiveness of the proposed method.

\bibliographystyle{splncs}
\bibliography{egbib}

\end{document}